\pdfoutput=1
\documentclass[11pt]{article}

\usepackage{emnlp2021}
\usepackage{times}
\usepackage{latexsym}

\usepackage[T1]{fontenc}
\usepackage[utf8]{inputenc}

\newcommand{\figref}[1]{Figure~\ref{#1}}
\newcommand{\tabref}[1]{Table~\ref{#1}}

\usepackage{booktabs}
\usepackage{threeparttable}
\usepackage{multirow}
\usepackage{color}

\usepackage{epsfig}
\usepackage{graphicx}
\usepackage{subfigure}
\usepackage{overpic}
\usepackage{amsfonts}
\usepackage{bm}
\usepackage{amsmath}
\usepackage{bbm}
\usepackage{url}

\usepackage{microtype}

\title{Learning to Ground Visual Objects for Visual Dialog}

\author{{Feilong Chen}\quad {Xiuyi Chen}\quad {Can Xu}\quad {Daxin Jiang}\thanks{\;\;Corresponding author.} \\
Microsoft Corporation, Beijing, China \\
  {\tt
  \{\href{mailto:ivess.chan@gmail.com}{ivess.chan},\href{mailto:hugheren.chan@gmail.com
}{hugheren.chan}\}@gmail.com}\\
  {\tt \{\href{mailto:caxu@microsoft.com}{caxu},\href{djiang@tencent.com}{djiang}\}@microsoft.com}\\
  }

\begin{document}
\maketitle
\begin{abstract}
Visual dialog is challenging since it needs to answer a series of coherent questions based on understanding the visual environment. How to ground related visual objects is one of the key problems. Previous studies utilize the question and history to attend to the image and achieve satisfactory performance, however these methods are not sufficient to locate related visual objects without any guidance. The inappropriate grounding of visual objects prohibits the performance of visual dialog models. In this paper, we propose a novel approach to Learn to Ground visual objects for visual dialog, which employs a novel visual objects grounding mechanism where both prior and posterior distributions over visual objects are used to facilitate visual objects grounding. Specifically, a posterior distribution over visual objects is inferred from both context (history and questions) and answers, and it ensures the appropriate grounding of visual objects during the training process. Meanwhile, a prior distribution, which is inferred from context only, is used to approximate the posterior distribution so that appropriate visual objects can be grounded even without answers during the inference process. Experimental results on the VisDial v0.9 and v1.0 datasets demonstrate that our approach improves the previous strong models in both generative and discriminative settings by a significant margin.
% \footnote{We will release the code at GitHub upon publication.}
\end{abstract}

% \begin{abstract}
% Visual dialog is challenging since it needs to answer a series of coherent questions based on understanding the visual environment. How to attend to related visual objects is one of the key problems. Previous studies utilize the question and history to attend to the image, while these methods are not sufficient to locate related visual objects. The inappropriate grounding of image objects prohibits the model from learning to make full use of the image. In this paper, we propose a novel model to Learn to Ground visual objects for visual dialog via Answer-Aware Knowledge Distillation (LG), which is achieved using a teacher-student framework. The teacher learns the posterior distribution of visual objects, that is, given the question, the history, the image, and the ground-truth answer to learn how to ground answer-aware visual objects from these four information. The student learns the prior distribution, that is, it learns to reconstruct the answer-aware visual object grounding only from the first three information. We transfer the ability of visual object grounding from the teacher to the student by leveraging knowledge distillation to close the gap between the two distributions. Experimental results on the VisDial v0.9 and v1.0 datasets demonstrate that our model improves the previous state-of-the-art visual dialog models in both generative and discriminative settings and achieves the new state-of-the-art performance in generative settings.
% \end{abstract}

\section{Introduction}

\begin{figure}[t!]
\centering
\scalebox{0.98}{
  \begin{overpic}[width=\columnwidth]{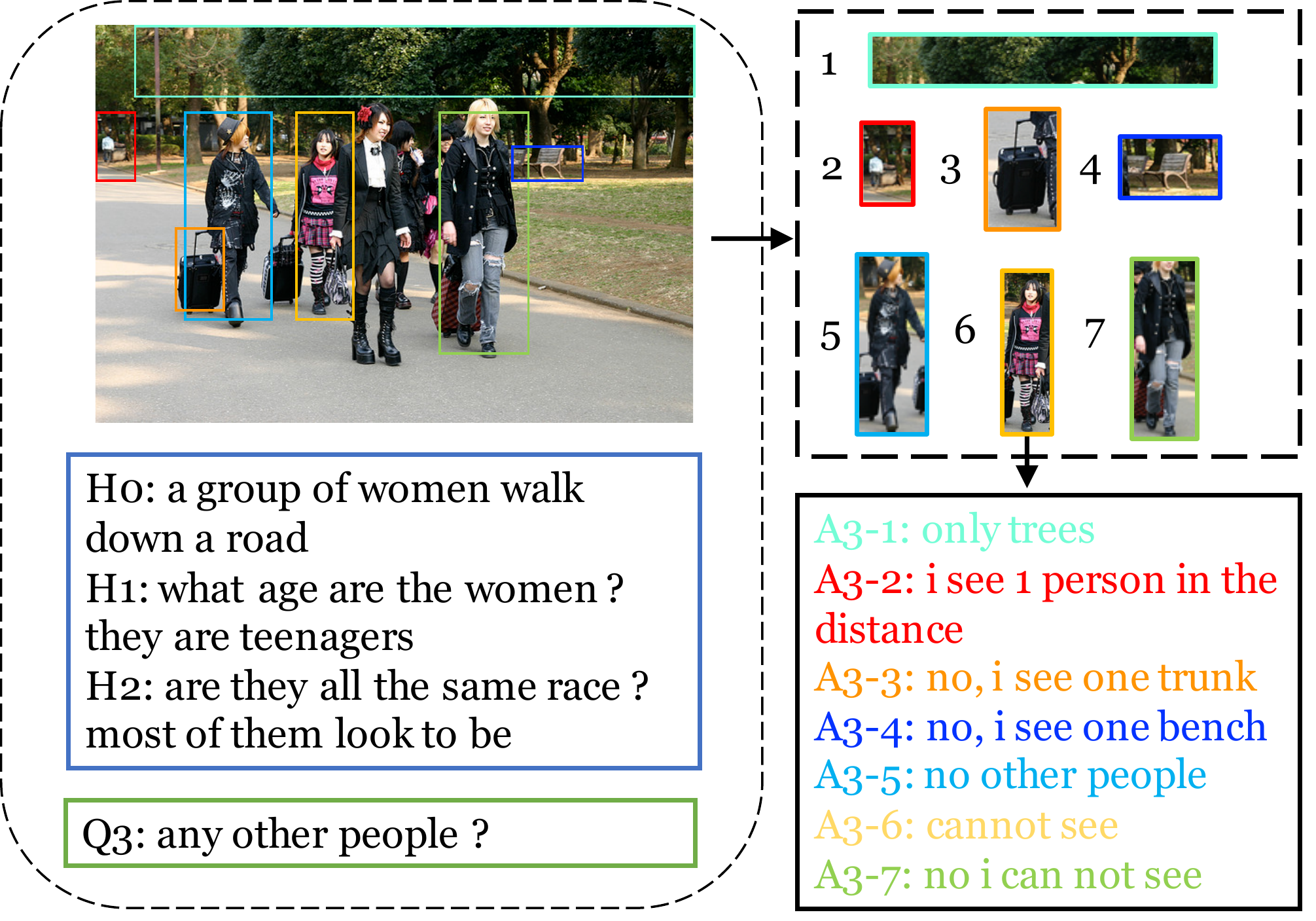}
  \end{overpic}
  }
  \caption{Comparison between different responses when focusing on different visual objects. We see that when the model focuses on wrong visual objects it makes mistakes. (Only the response A3-2 is right.) 
  }\label{fig:LG} 
\end{figure}

With the development of deep learning, various vision-language tasks have been introduced and attracted widespread attention, such as image captioning~\cite{xu2015show,Anderson2016SPICE,anderson2018bottom,cornia2020m2,ghanimifard2019goes}, visual question answering~\cite{ren2015exploring,gao2015you,lu2016hierarchical,anderson2018bottom,li2019relation,huang2020aligned} and visual dialog~\cite{das2017visual,Chen2021GoGRG,agarwal2020history,Chen2021MultimodalIT,Qi2020TwoCP}. Specifically, visual dialog, which aims to hold a meaningful conversation~\cite{Chen2021UnsupervisedKS,Chen2020BridgingTG} with a human about a given image, is a challenging task that requires models to locate related visual objects in an image and answer the current question based on the history and the located visual objects.

In order to answer the question correctly, we need to accurately locate the question-related visual objects. Most existing methods utilize kinds of attention mechanism~\cite{lu2017best,wu2018you,Kottur2018VisualCR,gan2019multi,guo2019dual} to capture the target visual objects. ReDAN~\cite{gan2019multi} and DMAM~\cite{chen2020dmrm} use multi-step reasoning based on dual attention to iteratively update related visual objects. DAN~\cite{guo2019dual}, MCAN~\cite{agarwal2020history} and LTMI~\cite{nguyenefficient} utilize multi-head attention mechanisms to manage multi-modal intersection and obtain weight distributions. Moreover, there are some approaches~\cite{zheng2019reasoning,schwartz2019factor,jiang2020dualvd,guo2020iterative,jiang2020kbgn} using graph-based structures to capture related visual objects. FGA~\cite{schwartz2019factor} realizes a factor graph attention mechanism, which constructs the graph over all the multi-modal features and estimates their interactions to ground visual objects. CAG~\cite{guo2020iterative} focuses on an iterative question-conditioned context-aware graph to locate related visual objects. However, the methods mentioned above obtain the prior distribution of visual objects through various interactions of questions, history and images, and finally use the prior distribution to obtain the final representation of the image. The prior distribution of visual objects is not enough to ground accurate visual objects, thus obtaining the wrong representation of the image.

In this paper, we propose a method to learn to ground visual objects in visual dialog. Specifically, we obtain the posterior distribution over visual objects by utilizing contexts and answers, while the prior distribution works without knowing answers in advance. Then we minimize the distance between the two distributions. During the training process, our model is trained to minimize the KL divergence between the prior distribution and the posterior distribution so that our model can approximate the posterior distribution accurately using the prior distribution. Then, during the inference process, the model grounds visual objects merely
based on the prior distribution (i.e., without any posterior information). We show that through this process, the model can effectively learn to ground visual objects accurately and give informative and accurate responses by utilizing appropriate visual objects. We test the effectiveness of our proposed model on two large-scale datasets: VisDial v0.9 and v1.0~\cite{das2017visual}. The contributions of this work are summarized as follows:
% In this paper, we propose a method to learn to ground visual objects in visual dialog based on a teacher-student framework. Specifically, the teacher is given the question, the history, the image, and the ground-truth answer to learn how to ground answer-aware visual objects, thus obtaining the posterior distribution of visual objects. The student learns to reconstruct the answer-anticipated visual object grounding only from the first three sources, thus obtaining the prior distribution of visual objects. We utilize knowledge distillation to make the prior distribution approximate the posterior distribution. During inference on testing data, the student
% will be applied. We test the effectiveness of our proposed model on two large-scale datasets: VisDial v0.9 and v1.0~\cite{das2017visual}. Both automatic and manual evaluations show that our approach can be used to improve the prior state-of-the-art models. The contributions of this work are summarized as follows:
\begin{itemize}
  \item We explore the importance of answers in grounding visual objects related to questions in visual dialog.
  \item We propose a novel approach to realize learning to ground visual objects in visual dialog via bridging the gap between the prior and posterior distribution over visual objects.
  \item We conduct extensive experiments and ablation studies on two large-scale datasets VisDial v0.9 and v1.0. Experimental results show that our approach can be used to improve previous visual dialog models in both generative and discriminative settings.
\end{itemize}

\begin{figure}[t]
\centering
\scalebox{0.98}{
  \begin{overpic}[width=\columnwidth]{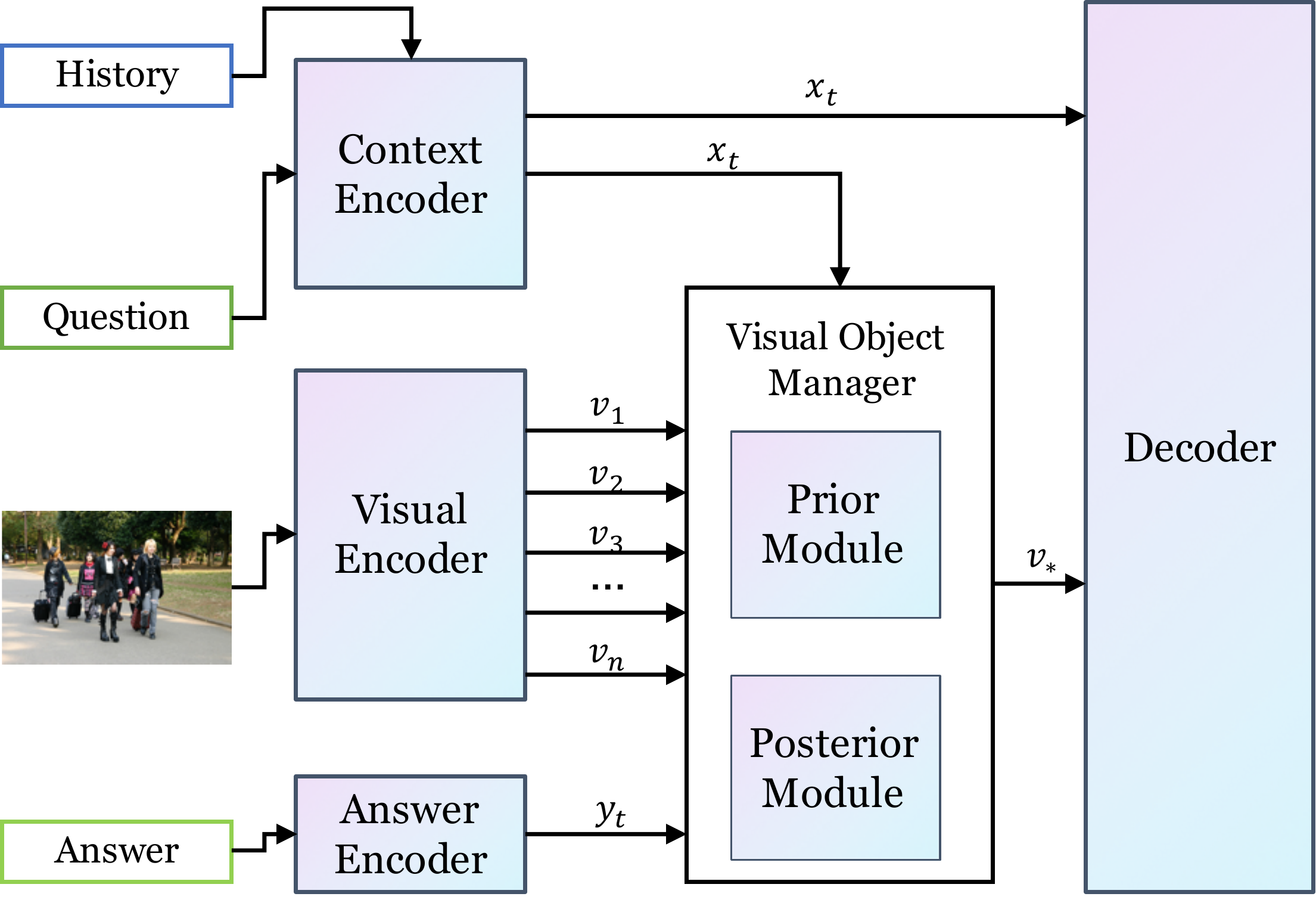}
  \end{overpic}
  }
  \caption{Architecture Overview
  }\label{fig:model_overview} 
\end{figure}

\begin{figure*}[t]
\centering
\scalebox{0.98}{
  \begin{overpic}[width=\textwidth]{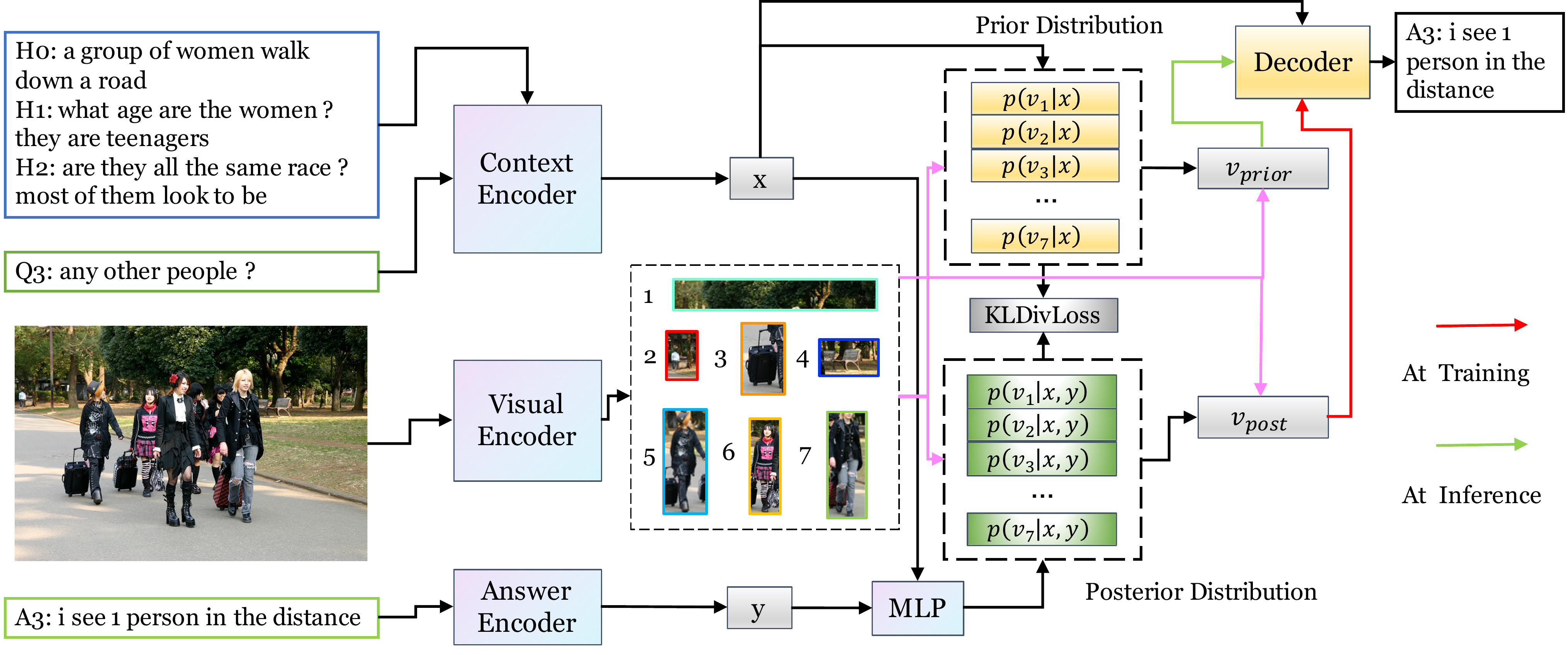}
  \end{overpic}
  }
  \caption{Framework of our Learning to Ground Visual Objects. The context encoder encodes the history and the current question into a context representation $x$. The visual encoder encodes the image into a region-based image features $v$. The answer encoder encodes the ground-truth answer into a response representation $y$. The visual object manager takes $x$, $v$ and $y$ as inputs, and generate a prior distribution $p(v|x)$ over visual objects and a posterior distribution $p(v|x,y)$ over visual objects, thus generating the new prior visual object features $v_{prior}$ and the posterior visual object representation $v_{post}$. The decoder utilizes the context representation $c$ and the new visual object representation ($v_{post}$ at training or $v_{prior}$ at inference) to generate and retrieve a response.
  }\label{fig:model} 
\end{figure*}

\section{Methodology}

According to \citeauthor{das2017visual}~\shortcite{das2017visual}, a visual dialog agent takes an image $i$, history (the caption and question-answer pairs) till round $t-1$: $h=(\underbrace{Cap}_{h_0}, \underbrace{(q_1, a_1)}_{h_1}, \cdots, \underbrace{(q_{t-1},a_{t-1})}_{h_{t-1}})$ and the current question $q_t$ at round $t$ as inputs. Note that $Cap$ is the caption describing the image taken as $h_0$ and $h_1, \dots, h_{t-1}$ are concatenations of question-answer pairs. Visual dialog agent aims to predict an answer $a_t$ to the question $q_t$. 
% To pinpoint accurate visual objects for question answering, we design a teacher-student framework to construct visual object distribution as follows:

% Following \citeauthor{das2017visual}~\shortcite{das2017visual}, a visual dialog agent is given three inputs, i.e., an image $i$, history (the caption and question-answer pairs) till round $t-1$: $h=(\underbrace{Cap}_{h_0}, \underbrace{(q_1, a_1)}_{h_1}, \cdots, \underbrace{(q_{t-1},a_{t-1})}_{h_{t-1}})$ and the current question $q_t$ at round $t$, where $Cap$ is the caption describing the image taken as $h_0$ and $h_1, \dots, h_{t-1}$ are concatenations of question-answer pairs. The goal of the visual dialog agent is to generate an answer $a_t$ to the question $q_t$.

\subsection{Model Architecture}
In this paper, we focus on training a neural visual dialog model with an effective visual objects grounding mechanism. As shown in~\figref{fig:model_overview}, we simplify existing visual dialog models into five major components:
\begin{itemize}
    \item {\bf The context encoder} encodes dialog history $h$ and the current question $q_t$ with an attention mechanism into a context vector $x$, and feeds it into the visual objects manager and the decoder.
    \item {\bf The visual encoder} takes the image $i$ as input and extract the image features $v = \{v_1, v_2, \dots, v_{\mu}\}$ where $\mu$ denotes the number of object proposals for each image. Each object proposal is represented by a $d_v$-dimension feature vector.
    \item {\bf The answer encoder} encodes the ground-truth answer $a_t$ into a response vector $y$, and feeds it into the visual objects manager.
    \item {\bf The visual object manager} consists of two sub-modules: a prior module and a posterior module. Given the previously encoded $x$ and ${v}_{i=1}^{\mu}$ (and $y$ if available), the visual object manager is responsible for deciding an appropriate distribution over visual objects and feeds the weighted visual object features $v_*$ (together with an attention-based context vector $x$) into the decoder.
    \item {\bf The decoder} generates and retrieves responses based on the visual object feature $v_*$ and the attention-based context vector $x$.
\end{itemize}

\subsection{Our Approach}
When given the context vector $x$ and the visual object features $v = \{v_1. v_2, \dots, v_{\mu}\}$, and response vector $y$, the goal of the visual object manager is to decide an appropriate distribution  $D$ over visual objects and obtain the weighted visual object representation $v_*$ based on the distribution $D$.

The visual object manager consists of two sub-modules: a prior module and a posterior module. 
\paragraph{The Prior Module.} 
The prior module aims to calculate the conditional probability distribution over $\mu$ visual objects, denoted by $\bf p(v|x)$:
\begin{equation}
    p(v=v_i|x) = \frac{exp(f_{cv}(x, v_i))}{\sum_{j=1}^{\mu}exp(f_{cv}(x, v_j))},
\end{equation}
where $f_{cv}(\cdot,\cdot)$ denotes the interaction function of the context vector $x$ and the visual object features $v_i$. For example, $f_{cv}(\cdot,\cdot)$ can be the dot product, self-attention or other mechanisms to measure the association between $v_i$ and the context vector $x$.  A high association means that $v_i$
is relevant to $x$ and thus, $v_i$
has a larger weight. Note that $p(v|x)$ is conditioned only on $x$ and thus, it is a prior distribution over visual objects since it works without knowing the response. However, there can be different visual objects that are relevant
to the contexts, and thus, it is difficult to select visual objects simply based on the prior distribution in training.

\paragraph{The Posterior Module.}
Motivated by this, in the posterior module, we define a posterior distribution over visual objects, denoted by $p(v|x, y)$, by considering both contexts and responses:
\begin{equation}
    p(v=v_i|x,y) = \frac{exp(f_{cv}(f_{cy}(x,y), v_i))}{\sum_{j=1}^{\mu}exp(f_{cv}(f_{cy}(x,y), v_j))},
\end{equation}
where $f_{cy}(\cdot,\cdot)$ denotes the interaction function of $x$ and $y$. For example, the  $f_{cy}(\cdot,\cdot)$ can be an add operation, fully connected layer and other methods. Compared with the prior distribution, the posterior distribution is sharp since the actual visual objects used in the true response $a_t$ can be captured.

\paragraph{Bridging the Gap.} 
Clearly, the discrepancy between prior and posterior distributions introduces great challenges in training the model: it is desirable to ground visual objects based on the posterior distribution, which, however, is unknown during inference. In
this paper, we propose to approximate the posterior distribution using the prior distribution so that our model is capable of selecting appropriate visual objects even without posterior information. For this purpose, we introduce an auxiliary loss, namely the Kullback-Leibler divergence loss (KLDivLoss), to bridge the gap between the prior distribution and the posterior distribution. The KLDivLoss is defined as follows:
\begin{equation}
    \mathcal{L}_{KL} = \sum_{i=1}^{\mu}p(v=v_i|x,y)log(\frac{p(v=v_i|x,y)}{p(v=v_i|x)}).
\end{equation}
When minimizing KLDivLoss, the posterior distribution $p(v|x,y)$ can be regarded as labels and our model is instructed to use the prior distribution $p(v|x)$ to approximate
$p(v|x, y)$ accurately. As a consequence, even when the posterior distribution is unknown in the inference process (since the actual response $a_t$ is unknown), the prior distribution $p(v|x)$ can be effectively utilized to ground appropriate
visual objects so as to generate and retrieve proper responses. To the best of our knowledge, it is the first neural model in visual dialog, which incorporates the posterior distribution as guidance, enabling accurate visual object grounding and high-quality response generation and retrieval.

\begin{table*}
\centering
\resizebox{0.97 \textwidth}!{
\begin{tabular}{lccccc|cccccc}
\toprule
\multirow{2}{*}{Model} &
\multicolumn{5}{c}{VisDial v0.9 (val)} &
\multicolumn{6}{c}{VisDial v1.0 (val)} \\
\cline{2-6}   \cline{7-12}
& MRR $\uparrow$ & R@1 $\uparrow$ & R@5 $\uparrow$ & R@10 $\uparrow$ & Mean $\downarrow$ & NDCG $\uparrow$ & MRR $\uparrow$ & R@1 $\uparrow$ & R@5 $\uparrow$ & R@10 $\uparrow$ & Mean $\downarrow$  \\
\midrule
% \multicolumn{12}{c}{Fusion-based Models}\\
% \midrule
LF~\cite{das2017visual} & 51.99 & 41.83 & 61.78 & 67.59 & 17.07
& - & - & - & - & - & -\\
HRE~\cite{das2017visual} & 52.37	& 42.23 & 62.28	& 68.11	& 16.97 
& - & - & - & - & - & -\\
% \midrule
MN~\cite{das2017visual}  & 52.59 & 42.29 & 62.85 & 68.88 & 17.06
& 51.86 & 47.99 & 38.18 & 57.54 & 64.32 & 18.60\\
HCIAE~\cite{lu2017best}  & 53.86 & 44.06 & 63.55 & 69.24 & 16.01 
& 59.70 & 49.07 & 39.72 & 58.23 & 64.73 & 18.43 \\
CorefNMN~\cite{Kottur2018VisualCR} & 53.50 & 43.66 & 63.54 & 69.93 & 15.69 
& - & - & - & - & - & - \\
CoAtt~\cite{wu2018you} & 54.11 & 44.32 & 63.82 & 69.75 & 16.47 
& 59.24 & 49.64 & 40.09 & 59.37 & 65.92 & 17.86 \\
RvA~\cite{niu2019recursive} & 55.43 & 45.37 & 65.27 & \underline{72.97} & \underline{10.71}
& - & - & - & - & - & - \\
DVAN~\cite{guo2019dual} & 55.94 & \underline{46.58} & 65.50 & 71.25 & 14.79
& - & - & - & - & - & - \\
Primary~\cite{guo2019image} & - & - & - & - & -
& - & 49.01 & 38.54 & 59.82 & 66.94 & 16.60 \\
ReDAN~\cite{gan2019multi} & - & - & - & - & -
& 60.47 & 50.02 & 40.27 & 59.93 & 66.78 & 17.40 \\
DMRM~\cite{chen2020dmrm} & \underline{55.96} & 46.20 & \underline{66.02} & 72.43 & 13.15
& - & 50.16 & 40.15 & 60.02 & 67.21 & 15.19 \\
DAM~\cite{jiang2020dam} & - & - & - & - & -
& 60.93 & \underline{50.51} & \underline{40.53} & \underline{60.84} & 67.94 & 16.65 \\
VDBERT~\cite{wang2020vd}$^\diamond$ & 55.95 & 46.83 & 65.43 & 72.05 & 13.18
& - & - & - & - & - & - \\
KBGN~\cite{jiang2020kbgn} & - & - & - & - & -
& 60.42 & 50.05 & 40.40 & 60.11 & 66.82 & 17.54 \\
\midrule
% LF~\cite{das2017visual}$^\dagger$ & 51.99 & 41.83 & 61.78 & 67.59 & 17.07
%  & - & - & - & - & - & -\\
% HCIAE~\cite{lu2017best}$^\dagger$ & 53.86 & 44.06 & 63.55 & 69.24 & 16.01 
%  & 59.70 & 49.07 & 39.72 & 58.23 & 64.73 & 18.43 \\
LTMI~\cite{nguyenefficient}$^\dagger$ & 55.85 & 46.07 & 65.97 & 72.44 & 14.17
& \underline{61.61} & 50.38 & 40.30 & 60.72 & \underline{68.44} & \underline{15.73} \\
% LTMI-Multi~\cite{nguyenefficient}$^\dagger$ & - & - & - & - & - 
% & 63.58 & 50.74 & 40.44 & 61.67 & 69.71 & 14.93 \\
% \midrule
% LF-LG (Ours) & - & - & - & - & - 
%  & - & - & - & - & - & -\\
% HCIAE-LG (Ours) & - & - & - & - & - 
%  & - & - & - & - & - & -\\
LTMI-LG (Ours) & 56.56 & 46.71 & 66.69 & 73.37 & 13.62 
& 63.23 & 51.30 & 41.34 & 61.61 & 69.06 & 15.26 \\
LTMI-LG$^*$ (Ours) & \bf{56.59} & \bf{46.87} & \bf{66.92} & \bf{73.76} & 13.35 
 & \bf{63.53} & \bf{51.43} & \bf{41.68} & \bf{61.96} & \bf{69.87} & \bf{14.89}\\
\bottomrule
\end{tabular}}
\caption{Main comparisons on both VisDial v0.9 and v1.0 datasets using the generative decoder. $\dagger$ denotes that we re-implemented the model using the released code. $\diamond$ denotes that the model utilizes large extra datasets for training which is unfair compared with other models. $*$ denotes that we train the model using multi-task learning. Underline indicates the highest performance among previous approaches except for pretraining-based models. Our approach improves the strong baseline a lot. (t-test, p-value$ \textless$0.01)} 
\label{tab:gen}
\end{table*}

\section{Application of Our Approach}
We take the strong baseline LTMI~\cite{nguyenefficient} as a base model to introduce our approach, which mainly consists of the following components:
\paragraph{Context Encoder and Answer Encoder:}We use two bi-directional LSTM
encoders to extract token-level representations ${\bf Q} \in \mathbb{R}^{\lambda \times d_q}$ and ${\bf y} \in \mathbb{R}^{\lambda \times d_q}$ of the question $q_t$ and the answer $a_t$. We use another bi-directional LSTM encoder to extract sentence-level representations ${\bf H} \in \mathbb{R}^{T \times d_q}$ of the history $h$. $\lambda$ is the length of questions and answers with paddings, $T$ is the turn of dialog and $d_q$ is the dimension. ${\bf Q}$ and ${\bf H}$ are fused into a context representation $\bf x$ with multi-head attention~\cite{vaswani2017attention}.
\paragraph{Visual Encoder:} Similar to~\cite{anderson2018bottom}, we extract the image features by using a pre-trained Faster RCNN~\cite{ren2015faster}. We select $\mu$ object proposals for each image, where each object proposal is represented by a 2048-dimension feature vector. We transform the obtained visual region features by a multi-layer perceptron and obtain the image features ${\bf I}={{\bf I}_{I=0}^{\mu}} \in \mathbb{R}^{\mu \times d_q}$.
\paragraph{Prior Module:} We use multi-head attention~\cite{vaswani2017attention} as $f_{cv}(\cdot, \cdot)$ to manage the multi-modal interaction. A cross-attention layer is firstly applied to outputs of the texutal and visual encoders:
\begin{eqnarray}
    {\bf P} &=& {\rm softmax}({\bf I}{\bf x}^T) \in \mathbb{R}^{\mu \times \lambda},\\
    {\bf I}_x &=& {\rm CrossAttn}({\bf I}, {\bf x})={\bf P}{\bf x} \in \mathbb{R}^{\mu \times d_q},
\end{eqnarray}
where the softmax conducts the normalization over each column of the matrix.
We convert the representation ${\hat {\bf I}}$ into $d_q$-dimension vectors ${\bf V}$. This conversion is performed by a simple self-attention computation as follows:
\begin{equation}
    {\bf g} = {\rm softmax}({\rm ReLU}({\hat {\bf I}}{\bf W}_1 + {\bf b}_1){\bf W}_2 + {\bf b}_2), \label{eq:fusion1}
\end{equation}
where ${\bf g} \in \mathbb{R}^{\mu \times 1}$, ${\bf W}_1$, ${\bf W}_2$, ${\bf b}_1$, ${\bf b}_2$ are learned parameters. We obtain the representation $\bf V$ as follows:
\begin{equation}
    {\bf v}_{prior} = {\bf g}^T{\hat {\bf I}} \in \mathbb{R}^{d_q}. \label{eq:fusion2}
\end{equation}
$\bf g$ is regarded as the prior distribution over visual objects.
\paragraph{Posterior Module:} We simply utilize the add operation as $f_{cy}(\cdot, \cdot)$ to manage the interaction of $\bf x$ and $\bf y$:
\begin{equation}
    {\bf x}_y = {\bf x} + {\bf y}
\end{equation}
We replace $\bf x$ in Eq.(6) - Eq.(8) with ${\bf x}_y$ and thus obtain the posterior distribution $\bf G$ and ${\bf v}_{post}$

\paragraph{Generative and Discriminative Decoder:} 
We utilize another LSTM as our discriminative and generative decoders following the previous studies~\cite{das2017visual,nguyenefficient}. Receiving the representation of context, images and the candidate answers, the two decoders compute the score of each candidate answer in different ways. The objective function of the base model is to minimize the negative log-likelihood $\mathcal{L_G}$ of answer generated for the generative decoder or the cross-entropy loss $\mathcal{L_D}$ for the discriminative decoder. We utilize the Kullback-Leibler (KL) divergence loss to narrow the gap. The objective functions of the student are as follows:
\begin{eqnarray}
    \mathcal{L} &=& \mathcal{L_G} + \lambda\mathcal{L}_{KL}({\bf G}, {\bf g}), \\
    \mathcal{L} &=& \mathcal{L_D} + \lambda\mathcal{L}_{KL}({\bf G}, {\bf g}), \\
    \mathcal{L} &=& \mathcal{L_G}+ \mathcal{L_D} + \lambda\mathcal{L}_{KL}({\bf G}, {\bf g}), 
\end{eqnarray}

\section{Experiments}

\subsection{Experiment Setup}
\paragraph{Datasets.}
Our experiments are conducted on the VisDial v0.9 and v1.0 datasets~\cite{das2017visual}. VisDial v0.9 contains 1.23M dialog question-answer pairs totally with 83k dialog for training and 40k dialogs for validation. VisDial v1.0 dataset is an extension of VisDial v0.9 dataset, which contains 123k, 2k, and 8k images as train, validation, and test splits, respectively.

\paragraph{Implementation Details.}
A Faster R-CNN~\cite{ren2015faster} with ResNet-101~\cite{he2016deep} finetuned on the Visual Genome dataset~\cite{krishna2017visual} are used to represent image regions. We set $\mu$ = 100 regions from each image and we use a 2048-dimension feature vector for each region. In experiments, we use Adam~\cite{kingma2014adam} optimizer for training, with the mini-batch size as 32. For the choice of the learning rate, we employ the warm-up strategy~\cite{goyal2017accurate}. Specifically, we begin with a learning rate of 0.001, the learning rate is decreased by 1/4 for every 2 epochs up to 20 epochs. We use 4 Titan-XP GPU for training. We spend about 4 hour / 1 epoch for the discriminative setting and 1 hour / 1 epoch for the generative setting. Our student model is as the same as LTMI, with the total parameters 42.20M. The $\lambda$ sets to 1.

\paragraph{Automatic Evaluation.}
To be consistent with previous work~\cite{das2017visual,chen2021gog,chen2021multimodal}, we use a retrieval setting to evaluate individual responses at each round of a dialog. Specifically,  there is a list of 100-candidate answers to be given at test time. We evaluate visual dialog models on retrieval metrics: (1) Rank of human response (the lower the better), (2) Existence of the human response in $top-k$ ranked responses, i.e., R@$k$ (the higher the better) (3) Mean Reciprocal Rank (MRR) of the human response (the higher the better) and (4) Normalized Discounted Cumulative Gain (NDCG, the higher the better) for VisDial v1.0.

\paragraph{Human Evaluation.}
Following~\cite{wu2018you,chen2021gog}, we randomly extract 100 predicted samples for human evaluation. We ask 3 human subjects to guess whether the last response in the dialog is human-generated or machine-generated. If at least 2 of them agree it is generated by a human, we think it passes the Truing Test (M1). We record the percentage of responses that are evaluated better than or equal to human responses (M2), according to the human subjects’ evaluation.

\begin{table*}
\centering
\resizebox{0.99\textwidth}!{
\begin{tabular}{lccccc|cccccc}
\toprule
\multirow{2}{*}{Model} &
\multicolumn{5}{c}{VisDial v0.9 (val)} &
\multicolumn{6}{c}{VisDial v1.0 (test-std)} \\
\cline{2-6}   \cline{7-12}
& MRR $\uparrow$ & R@1 $\uparrow$ & R@5 $\uparrow$ & R@10 $\uparrow$ & Mean $\downarrow$ & NDCG $\uparrow$ & MRR $\uparrow$ & R@1 $\uparrow$ & R@5 $\uparrow$ & R@10 $\uparrow$ & Mean $\downarrow$  \\
\midrule
% \multicolumn{12}{c}{Fusion-based Models}\\
% \midrule
LF~\cite{das2017visual} & 58.07 & 43.82 & 74.68 & 84.07 & 5.78
& 45.31 & 55.42 & 40.95 & 72.45 & 82.83 & 5.95\\
HRE~\cite{das2017visual} & 58.46 & 44.67 & 74.50 & 84.22	& 5.72 
& 45.46 & 54.16 & 39.93 & 70.45 & 81.50 & 6.41\\
% \midrule
% \multicolumn{12}{c}{Attention-based Model} \\
% \midrule
MN~\cite{das2017visual}$^\dag$ & 59.65 & 45.55 & 76.22 & 85.37 & 5.46
& 47.50 & 55.49 & 40.98 & 72.30 & 83.30 & 5.92\\
HCIAE~\cite{lu2017best}$^\dag$ & 62.22 & 48.48 & 78.75 & 87.59 & 4.81 
& - & - & - & - & - & - \\
AMEM~\cite{seo2017visual} & 62.27 & 48.53 & 78.66 & 87.59 & 4.86 
& - & - & - & - & - & - \\
% to do
CoAtt~\cite{wu2018you} & 63.98 & 50.29 & 80.18 & 88.81 & 4.47 
& - & - & - & - & - & - \\
CorefNMN~\cite{Kottur2018VisualCR} & 64.10 & 50.92 & 80.18 & 88.81 & 4.45 
& 54.70 & 61.50 & 47.55 & 78.10 & 88.80 & 4.40 \\
RvA~\cite{niu2019recursive} & 66.34 & 52.71 & 82.97 & 90.73 & 3.93
& 55.59 & 63.04 & 49.03 & 80.40 & 89.83 & 4.18 \\
DVAN~\cite{guo2019dual} & 66.67 & 53.62 & 82.85 & 90.72 & 3.93
& 54.70 & 62.58 & 48.90 & 79.35 & 89.03 & 4.36 \\
Primary~\cite{guo2019image} & - & - & - & - & -
& 57.32 & 62.20 & 47.90 & 80.43 & 89.95 & 4.17 \\
ReDAN~\cite{gan2019multi} & - & - & - & - & -
& 57.63 & 64.75 & 51.10  & 81.73  & 90.90  &  3.89  \\
% DAN~\cite{kang2019dual} & 66.38 & 53.33 & 82.43 & 90.38 & 4.04
% & 57.59 & 63.20 & 49.63 & 79.75 & 89.35 & 4.30 \\
% HACAN~\cite{Yang2019ICCV} & 67.92 & 54.76 & 83.03 & 90.68 & 3.97 
% & 57.17 & 64.22 & 50.88 & 80.63 & 89.45 & 4.20 \\
MCA~\cite{agarwal2020history} & - & - & - & - & -
& \underline{72.73} & 37.68 & 20.67 & 56.67 & 72.12 & 8.89 \\
% \midrule
% \multicolumn{12}{c}{Pretraining-based Model} \\
% \midrule

% \midrule
% \multicolumn{12}{c}{Graph-based Model} \\
% \midrule
GNN-EM~\cite{zheng2019reasoning} & 62.85 & 48.95 & 79.65 & 88.36 & 4.57
& 52.82 & 61.37 & 47.33 & 77.98 & 87.83 & 4.57 \\
DualVD~\cite{jiang2020dualvd} & 62.94 & 48.64 & 80.89 & 89.94 & 4.17
& 56.32 & 63.23 & 49.25 & 80.23 & 89.70 & 4.11 \\
FGA~\cite{schwartz2019factor} & 65.25 & 51.43 & 82.08 & 89.56 & 4.35
& 56.90 & \underline{66.20} & \underline{52.75} &  \underline{82.92} & \underline{91.07} & \underline{3.80} \\
CAG~\cite{guo2020iterative} & \underline{67.56} & \underline{54.64} & \underline{83.72} & \underline{91.48} & \underline{3.75} 
& 56.64 & 63.49 & 49.85 & 80.63 & 90.15 & 4.11 \\
KBGN~\cite{jiang2020kbgn} & - & - & - & - & -
& 57.60 &  64.13 &  50.47 & 80.70 &  90.16 &  4.08 \\
\midrule
VisualBERT~\cite{murahari2019large}$^\diamond$ & - & - & - & - & - 
& 74.47 & 50.74 & 37.95 & 64.13 & 80.00 & 6.28 \\
VDBERT~\cite{wang2020vd}$^\diamond$ & 70.04 & 57.79 & 85.34 & 92.68 & 4.04
& 75.35 & 51.17 & 38.90 & 62.82 & 77.98 & 6.69 \\
\midrule
% LF~\cite{das2017visual}$^\dagger$ & 51.99 & 41.83 & 61.78 & 67.59 & 17.07
%  & - & - & - & - & - & -\\
% HCIAE~\cite{lu2017best}$^\dagger$ & - & - & - & - & -
% & - & - & - & - & - & - \\
% LTMI~\cite{nguyenefficient}$^\dagger$ & 66.41 & 53.36 & 82.53 & 90.54 & 4.03
% & 60.74 & 61.20 & 47.08 & 77.78 & 87.60 & 4.88 \\
LTMI~\cite{nguyenefficient}$^\dagger$ & 66.41 & 53.36 & 82.53 & 90.54 & 4.03
&60.92 & 60.65 & 47.00 & 77.03 & 87.75 & 4.90 \\
% \midrule
% LF-LG (Ours) & - & - & - & - & - 
%  & - & - & - & - & - & -\\
% HCIAE-LG (Ours) & - & - & - & - & - 
%  & - & - & - & - & - & -\\
% LTMI-LG (Ours) & - & - & - & - & - 
% & \bf{63.03} & \bf{51.14} & \bf{40.91} & \bf{61.78} & \bf{69.43} & \bf{15.08} \\
LTMI-LG (Ours) &  67.63  &  54.69  &  83.74  & 91.38 & \bf{3.75}
 & 58.55 & 64.00 & 50.63 & 80.58 & 90.20 & 4.12\\
\bottomrule
\end{tabular}}
\caption{Main comparisons on both VisDial v0.9 and v1.0 datasets using the discriminative decoder. $\diamond$ denotes that the model utilizes large extra datasets for training which is unfair compared with other models. Underline indicates the highest performance among previous approaches except for the pretraining-based models. Our approach improves the strong baseline significantly. (t-test, p-value$ \textless$0.01)}
\label{tab:disc_test}
\end{table*}

\subsection{Main Results}
\paragraph{Baseline methods.}
In our experiment, compared methods can be grouped into four types: (1) Fusion-based models: LF~\cite{das2017visual} and HREA~\cite{das2017visual}. (2) Attention-based models: HCIAE~\cite{lu2017best}, CoAtt~\cite{wu2018you}, Primary~\cite{guo2019image}, ReDAN~\cite{gan2019multi}, CorefNMN~\cite{Kottur2018VisualCR}, RvA~\cite{niu2019recursive}, DVAN~\cite{guo2019dual} and DMRM~\cite{chen2020dmrm}, DAM~\cite{jiang2020dam}. (3) The pretraining model: VDBERT~\cite{wang2020vd} and VisualBERT~\cite{murahari2019large}. (4) Graph-based models: GNN~\cite{zheng2019reasoning}, DualVD~\cite{jiang2020dualvd}, FGA~\cite{schwartz2019factor},
KBGN~\cite{jiang2020kbgn}. 
% Please refer to Appendix \textcolor{red}{A.3} for more compared methods. 

We realize our model LTMI-LG which is based on the strong baseline LTMI~\cite{nguyenefficient}\footnote{We reproduce the result for LTMI by their official GitHub repo (https://github.com/davidnvq/visdial). We apply the default hyper-parameters as them.}. LTMI is a very strong model which achieves some the-state-of-the-art results. In general, our approach brings a large improvement to the strong baseline LTMI, which shows the effectiveness of our answer-aware knowledge distillation. We use t-test and analysis of variance (ANOVA) to analyze our model and LTMI. The p-values of these two analytical methods are all less than 0.01, indicating that the results are significantly different.

\begin{table}[t]
    \centering
    \resizebox{0.99\columnwidth}!{
    \begin{tabular}{lccc}
    \toprule
     Model & w/o Ans & w/ Ans & with LG \\
     \midrule
    LTMI~\cite{nguyenefficient} & 68.6 & 97.1 & 82.1\\
    Random & 3.6 & 3.6 & -\\
    Human & 96.7 & 99.3 & -\\
    \bottomrule
    \end{tabular}}
    \caption{Accuracy of visual grounding with and without knowing the answer. We randomly sample 1000 samples and ask human annotators to ground the three most likely objects from the image.}
    \label{tab:visualgrounding}
\end{table}

\begin{table}[t]
    \centering
    \resizebox{0.99\columnwidth}!{
    \begin{tabular}{lcccccc}
    \toprule
     Model & NDCG & MRR & R@1 & R@5 & R@10 & Mean \\
     \midrule
     LTMI & 61.61 & 50.38 & 40.30 & 60.72 & 68.44 & 15.73 \\
     LTMI-Mean& 56.66  & 43.64 & 32.59 & 54.66 & 62.91 & 17.59 \\
     LTMI-Random  & 56.89  & 43.79 & 33.01 &  54.47 & 62.76 & 17.76\\
     LTMI-LG  & 63.23 & 51.30 & 41.34 & 61.61 & 69.06 & 15.26 \\
     LTMI-Human  & 70.10 & 63.96 & 50.74 & 69.29 & 80.02 & 8.12 \\
     \bottomrule
    \end{tabular}}
    \caption{Effects of different visual objects distribution.}
    \label{tab:visual}
\end{table}

\begin{figure*}[t]
\centering
\scalebox{0.9}{
  \begin{overpic}[width=\textwidth]{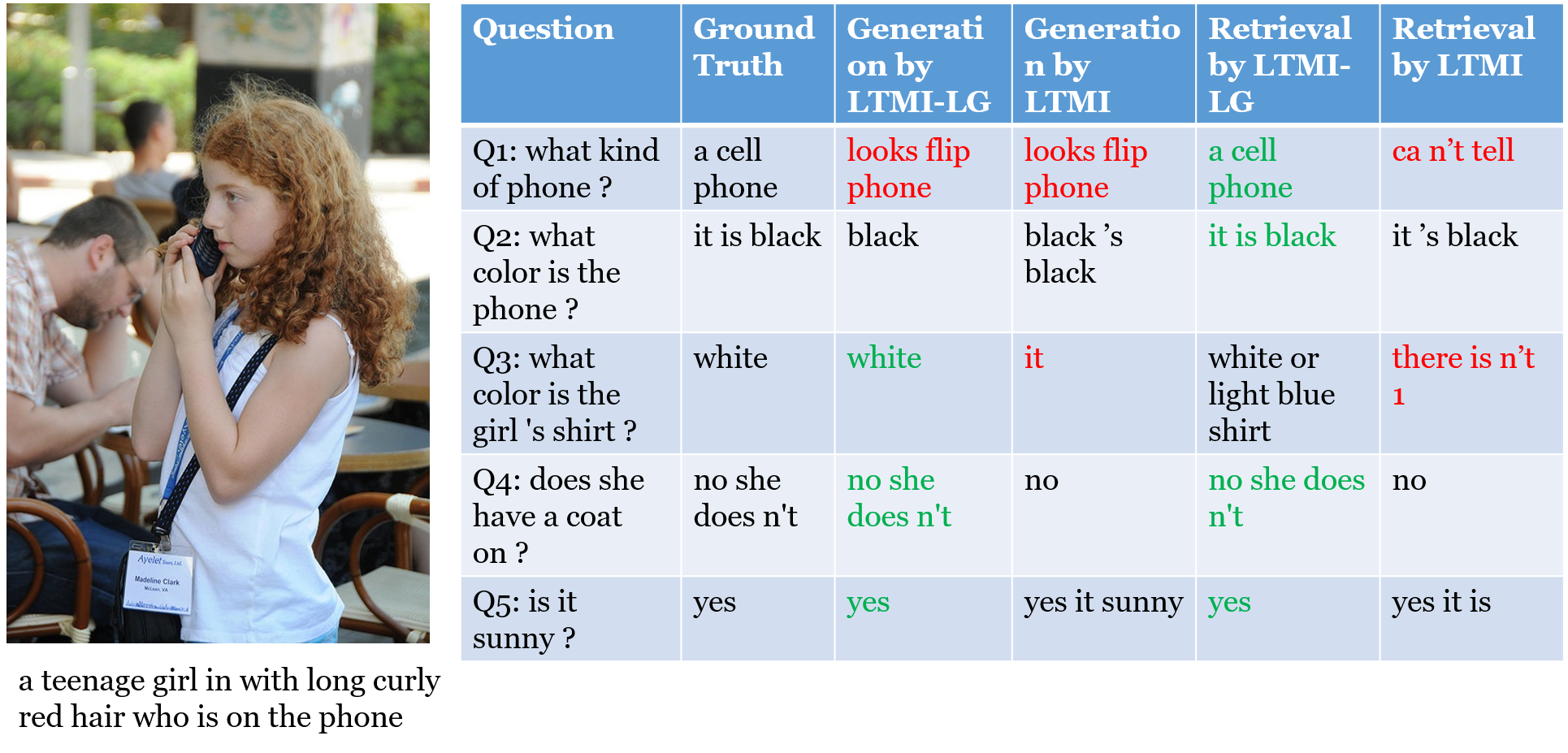}
  \end{overpic}
  } 
  \caption{Examples of dialogs generated and retrieved by our model and the LTMI baseline. Our model provides answers that are more accurate than LTMI (\textcolor{green}{green} denotes correct answers, and \textcolor{red}{red} denotes wrong answers). 
%   Results from our model are also more natural and comprehensive (highlighted in \textcolor{blue}{blue}).
  }\label{fig:dialog}
\end{figure*}

As shown in~\tabref{tab:visualgrounding}, we statistic the accuracy of grounding visual objects of our LTMI-LG, which is 82.1\%. Our answer-aware knowledge distillation improves the accuracy from 68.6\% (LTMI) to 82.1\% (LTMI-LG), gaining 13.5\% improvement. As shown in~\figref{fig:dialog}, we provide predicted answers by LTMI and our LTMI-LG. Due to the improvement of visual grounding, our approach improves the generative and retrieval results of LTMI, managing to locate visual objects more accurately, as shown in~\tabref{tab:visual}. ``Mean'' denotes We set the distribution of the visual objects to uniform to make all visual objects have the same weights. ``Random'' denotes we randomize the distribution in-batch. ``Human'' denotes we annotate 100 images and utilize this distribution to generate responses. The more appropriate visual objects, the better the model performance. 

\begin{figure*}[t]
\centering
\scalebox{0.85}{
  \begin{overpic}[width=\textwidth]{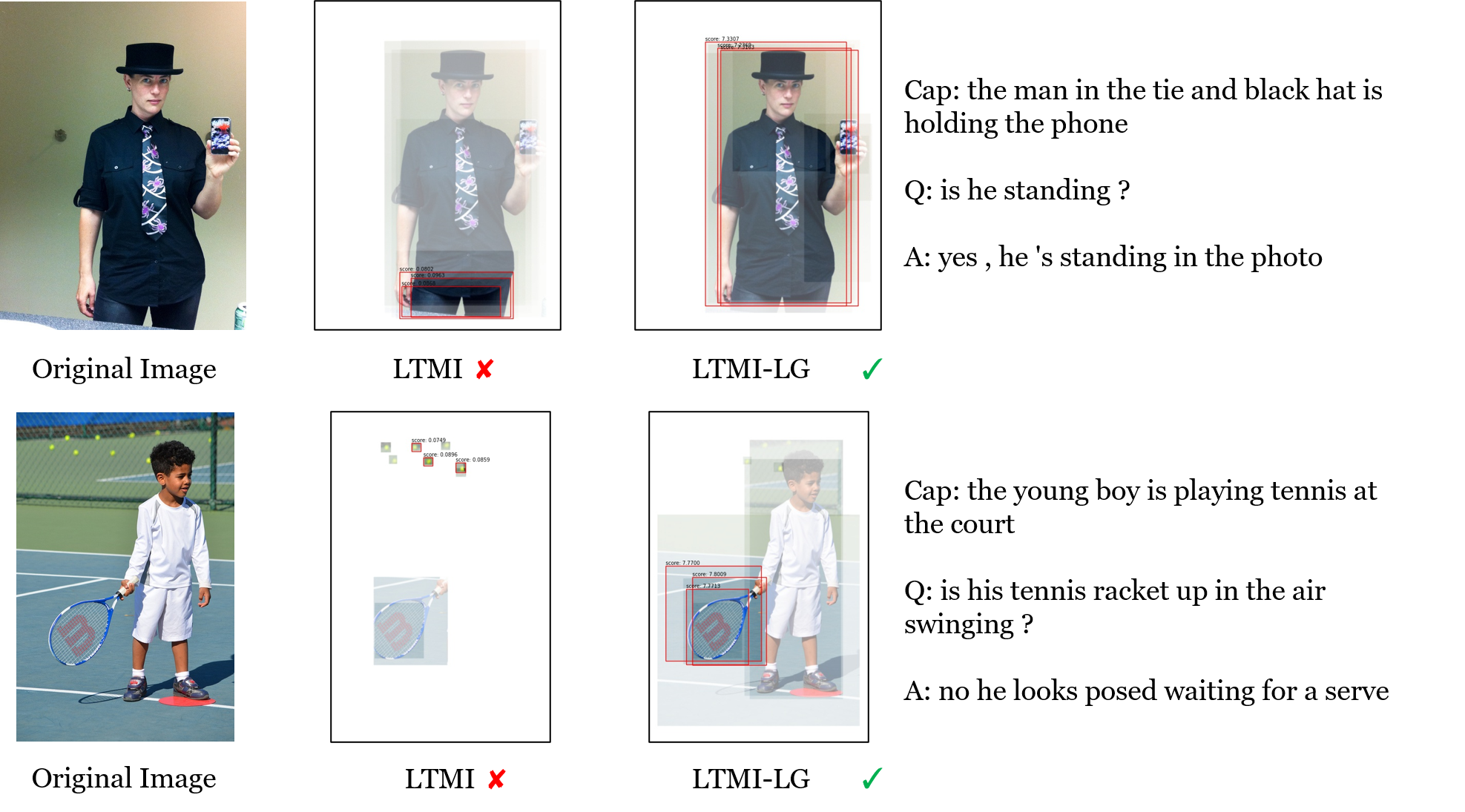}
  \end{overpic}
  }
  \caption{Visualization of attention maps generated by LTMI and our approach. Our approach grounds the related visual objects more accurately than LTMI.
  }\label{fig:example2}
\end{figure*}

\begin{table}[t]
    \centering
    \resizebox{0.99\columnwidth}!{
    \begin{tabular}{lcccccc}
    \toprule
     Model & NDCG & MRR & R@1 & R@5 & R@10 & Mean \\
     \midrule
     MN~\cite{das2017visual}& - & 60.29 & 46.14 & 77.68 & 87.57 & 4.84 \\
     HCIAE~\cite{lu2017best} & - & 61.96 & 48.25 & 78.97 & 88.43 & 4.56 \\
     CoAtt~\cite{wu2018you} & -  & 62.77 & 49.38 & 78.99 & 88.49 & 4.56 \\
     ReDAN~\cite{gan2019multi} & -  & 64.29 & 50.65 &  81.29 & 90.17 & 4.10 \\
     KBGN~\cite{jiang2020kbgn} & 59.08 & \underline{64.86} & \underline{51.37} & \underline{81.71} & \underline{90.54} & \underline{4.00}  \\
     VDBERT~\cite{wang2020vd}$^\ddag$ & 56.20 & 62.25 & 48.16 & 79.57 & 89.01 & 4.31 \\
     VDBERT~\cite{wang2020vd}$^\diamond$ & 63.22 & 67.44 & 54.02 & 83.96 & 92.33 & 3.53 \\
     \midrule
    %  HCIAE~\cite{lu2017best}$^\ddag$& 57.65 & 62.96 & 48.94 & 80.50 & 89.66 & 4.24 \\
    %  LTMI~\cite{nguyenefficient}$^\ddag$& 61.52  & 62.31 & 48.92 & 78.55 & 87.77 & 4.86 \\
     LTMI~\cite{nguyenefficient}$^\dagger$& \underline{62.72}  & 62.32 & 48.94 & 78.65 & 87.88 & 4.86 \\ 
    %  \midrule
    %  HCIAE-LG (Ours) & - & - & - & - & - & - \\
    %  LTMI-LG (Ours) & \bf{62.24}  & 63.81 & 50.33 &  80.48 & 89.24 & 4.35 \\
     LTMI-LG (Ours) & 59.67 &  65.03  &  51.69  & 81.49 & 90.32 & 4.02 \\
    \bottomrule
    \end{tabular}}
    \caption{Main comparisons on VisDial v1.0 val datasets using the discriminative decoder. $\diamond$ denotes that the model utilizes large extra datasets for training. $\ddag$ denotes that the model trains from scratch.}
    \label{tab:disc_val}
\end{table}

\paragraph{Generative Results.}
As shown in~\tabref{tab:gen}, we compare the generative performance among different methods on the VisDial v1.0 val and VisDial v0.9 val. With the guidance of the teacher, we train our LTMI-LG with the ability of accurately grounding visual objects. As a result, our approach improve significantly (nearly 1\% on all metrics) compared with LTMI~\cite{nguyenefficient}. Comparing with the state-of-the-art results of different metrics, our model improves NDCG for 61.51 to 63.53 (+1.92), MRR from 50.51 to 51.43 (+0.92), R@1 from 40.53 to 41.68 (+1.13), R@5 from 60.84 to 61.96 (+1.12), R@10 from 68.44 to 69.87 (+1.43), Mean from 15.73 to 14.89 (+0.84) on the VisDial v1.0 val. Our model also brings a large improvement to LTMI~\cite{nguyenefficient} on the VisDial v0.9 val. The performance of our model exceeds the performance of VDBERT~\cite{wang2020vd}$^\diamond$ on all the metrics except Mean. We believe data is an important factor in deep learning~\cite{lecun2015deep}. VDBERT~\cite{wang2020vd}$^\diamond$ works because it uses a lot of extra data for training. The reason why our method is effective is that we use the teacher to teach the student visual grounding, which can be regarded as a kind of data annotation.

\begin{table}[t]
    \centering
    \resizebox{0.99\columnwidth}!{
    \begin{tabular}{lcccccc}
    \toprule
     Model & NDCG & MRR & R@1 & R@5 & R@10 & Mean \\
     \midrule
    %  MN~\cite{das2017visual}& - & 60.29 & 46.14 & 77.68 & 87.57 & 4.84 \\
    LTMI$^\dagger$ & 61.61 & 50.38 & 40.30 & 60.72 & 68.44 &  15.73 \\
    % LTMI-Teacher & 63.19 & 51.01 & 40.87 & 61.62 & 68.97 & 15.28 \\
    \midrule
    LG-Attn-MSE & 63.03 & 51.14 & 40.91 & 61.78 & 69.43 & 15.08  \\
    LG-Image-MSE & 62.80 & 51.21 & 41.01 & \bf{62.02} & \bf{69.90} & \bf{14.90} \\
    % LG-Repre-MSE & 63.15 & 51.28 & 41.00 & \bf{62.07} & 69.80 & 14.91 \\
    \midrule
    % LTMI-LG-MSE & 63.03 & 51.14 & 40.91 & 61.78 & 69.43 & 15.08 \\
    % LTMI-LG-KL & 63.23 & 51.30 & 41.34 & 61.61 & 69.06 & 15.26 \\
    LG-Attn-KL & \bf{63.23} & \bf{51.30} & \bf{41.34} & 61.61 & 69.06 & 15.26 \\
    LG-Image-KL & 62.36 & 51.24& 41.19 & 61.73 & 69.31 & 15.13 \\
    % LG-Repre-KL & 62.68 & 51.25 & 41.22 & 61.57 & 69.01  & 15.29 \\
    \midrule
    Attn-KL-Image-MSE & 62.73 & 51.19 & 40.97 & 61.80 & 69.43 & 15.10 \\
    \bottomrule
    \end{tabular}}
    \caption{Ablation study on VisDial v1.0 val datasets using the generative decoder.}
    \label{tab:ablation}
\end{table}

\paragraph{Discriminative Results.}
As shown in~\tabref{tab:disc_test} and~\tabref{tab:disc_val}, we compare our method with previous works on the VisDial v1.0 test, VisDial v0.9 val and VisDial v1.0 val. Our model improves significantly compared with LTMI~\cite{nguyenefficient}, improving about +3\% on MRR, R@1, R@5 and R@10 on the VisDial v1.0 test. Our approach also brings a large improvement on the Visdial v0.9 and achieves the best results on MRR, R@1 and R@5 among non-pre-trained models. In a discriminative setting, our approach performs worse than pre-training models VisualBERT~\cite{murahari2019large}$^\diamond$ and VDBERT~\cite{wang2020vd}$^\diamond$ because pre-training models utilize extra large-scale datasets to train the models which are unfair compared with other models. As shown in~\tabref{tab:disc_val}, VDBERT$^\ddag$ which trains from scratch performs worse than our LTMI-LG.

\subsection{Ablation Study}
In order to transfer knowledge, we need a metric loss to measure the gap between teachers and students. In our main experiments, we utilize the kullback-leibler (KL) divergence loss to diminish the gap of the weight distribution between the student model and the teacher model, the mean squared loss to diminish the gap of the representation of images. To compare different losses, we utilize the mean squared loss for attention maps and KL loss for the representation of images as shown in~\tabref{tab:ablation}. We find that KL loss is more suitable for attention distribution (better for NDCG, MRR and R@1) and MSE loss for the representation (better for R@5, R@10 and Mean). In addition, we use the attention via KL loss and the representation via MSE loss for distillation at the same time. The result is not so satisfactory and we think these two methods have some redundancy.

\subsection{Case Study}
As shown in~\figref{fig:example2}, we visualize the learned attention maps to understand the model. The colorful region means higher attention weights. We draw the bounding boxes of the first three highest scores. As shown in the top image in \figref{fig:example2}, the question ``is he standing ?" indicates the man's overall posture rather than the local. LTMI grounds the wrong visual objects while our model grounds the right objects. As shown in the bottom image in \figref{fig:example2}, the question ``is his tennis racket up in the air swinging ?'' concerns the racket rather than the tennis balls. Our model grounds accurately while LTMI makes mistakes. These examples show that our LTMI-LG has learned the ability to ground visual objects via our answer-aware knowledge distillation. 

\subsection{Human Study}
As shown in~\tabref{tab:humanstudy}, we conduct human study to further demonstrate the effectiveness of our model. Our model achieves the highest scores both on the metric M1 and M2 compared with LTMI. 

\begin{table}
  \centering
%   \scalebox{0.85}{
    \resizebox{0.65\columnwidth}!{
  \begin{tabular}{p{2.5cm}|c|c}
    \toprule
      & LTMI & LTMI-LG\\
    \midrule
    Method 1 (M1) &  56 & 66\\
    \midrule
    Method 2 (M2) & 61 & 69\\
    \bottomrule
  \end{tabular}
  }\vspace{3pt}
  \caption{Human evaluation on 1000 sampled responses on VisDial val v1.0. 
  M1: percentage of responses which are human-generated. M2: percentage of responses evaluated better than or equal to human responses. 
  }\label{tab:humanstudy}
\end{table}

\section{Related Work}
Recent several works~\cite{shuster2018image,liang2021maria,yang2020open} explore leveraging visual information to enhance dialogue models. While visual dialog models focus on the intersection of questions, history and images. How to locate the related visual objects is quite important. MN~\cite{das2017visual}, HCIAE~\cite{lu2017best}, CorefNMN~\cite{Kottur2018VisualCR}, CoAtt~\cite{wu2018you}, RvA~\cite{niu2019recursive}, DVAN~\cite{guo2019dual} utilize kinds of attention mechanisms as the backbone to locate the related visual objects. VisualBERT~\cite{murahari2019large} and VDBERT~\cite{wang2020vd} exploit large extra datasets to explore in visual dialog via pretraining language models. GNN-EM~\cite{zheng2019reasoning}, FGA~\cite{schwartz2019factor}, DualVD~\cite{jiang2020dualvd}, CAG~\cite{guo2020iterative} and KBGN~\cite{jiang2020kbgn} utilize graph neural networks to obtain the representation of visual objects. However, most existing visual dialog models condition visual objects simply on history and questions, which we
regard as a prior distribution over visual objects. In this paper, we propose an approach to learn to ground visual objects via bridge the gap between the prior distribution and the posterior distribution.

% \paragraph{Knowledge Distillation}
% Knowledge distillation~\cite{hinton2015distilling,gou2020knowledge,mirzadeh2020improved,romero2014fitnets,liu2019knowledge,jiao2019tinybert} has been widely used in different fields of artificial intelligence, including visual recognition, speech, recognition, natural language processing, and recommendation systems.~\citeauthor{zagoruyko2016paying}~\shortcite{zagoruyko2016paying} improve the performance of a student CNN network by forcing it to mimic the attention maps of a powerful teacher network via properly defining attention for convolutional neural networks. \citeauthor{tian2020response}~\shortcite{tian2020response} propose a method to construct a response-anticipated memory to contain document information that is potentially more important in generating responses via knowledge distillation. In this paper, we propose an answer-aware knowledge distillation to teach the student model to learn to ground visual objects.

\section{Conclusion}
In this paper, we propose a novel approach to learn to ground visual objects for visual dialog, which employs a novel visual objects grounding mechanism where both prior and posterior distributions over visual objects are used to facilitate visual objects grounding. Experimental results on two large-scale datasets show that our approach improves the previous models by a significant margin.

% Entries for the entire Anthology, followed by custom entries
\bibliography{anthology,custom}
\bibliographystyle{acl_natbib}

% \appendix
% \newpage
% \section{Appendix}
% \label{sec:appendix}

% This is an appendix.

\end{document}